\documentclass[sigconf]{acmart}

\AtBeginDocument{%
  }

\copyrightyear{2025}
\acmYear{2025}
\setcopyright{acmlicensed}\acmConference[MM '25]{Proceedings of the 33rd ACM International Conference on Multimedia}{October 27--31, 2025}{Dublin, Ireland}
\acmBooktitle{Proceedings of the 33rd ACM International Conference on Multimedia (MM '25), October 27--31, 2025, Dublin, Ireland}
\acmDOI{10.1145/3746027.3755754}
\acmISBN{979-8-4007-2035-2/2025/10}

\RequirePackage[font=footnotesize,skip=3pt,subrefformat=parens]{subcaption}
\usepackage[capitalize]{cleveref}
\usepackage{colortbl}
\usepackage{pifont}
\definecolor{myblue}{RGB}{218,232,252}

\usepackage{xcolor}
\usepackage{soul}

\acmSubmissionID{5738}

\begin{document}

\title{Input Domain Aware MoE: Decoupling Routing Decisions from Task Optimization in Mixture of Experts}

\author{YongXiang Hua}
\orcid{https://orcid.org/0009-0008-8849-4717}
\affiliation{%
  \institution{University of Science and Technology of China}
  \institution{State Key Laboratory of Cognitive Intelligence}
  \city{Hefei}
  \country{China}}
\email{yx15333063290@mail.ustc.edu.cn}

\author{Haoyu Cao}
\orcid{https://orcid.org/0000-0002-3789-9705}
\affiliation{%
  \institution{University of Science and Technology of China}
  \institution{State Key Laboratory of Cognitive Intelligence}
  \city{Hefei}
  \country{China}}
\email{caohaoyu@mail.ustc.edu.cn}

\author{Zhou Tao}
\orcid{https://orcid.org/0009-0005-4433-4373}
\affiliation{%
  \institution{University of Science and Technology of China}
  \institution{State Key Laboratory of Cognitive Intelligence}
  \city{Hefei}
  \country{China}}
\email{zhoutao24@mail.ustc.edu.cn}

\author{Bocheng Li}
\orcid{https://orcid.org/0009-0005-2171-7902}
\affiliation{%
  \institution{University of Science and Technology of China}
  \institution{State Key Laboratory of Cognitive Intelligence}
  \city{Hefei}
  \country{China}}
\email{bcli@mail.ustc.edu.cn}

\author{Zihao Wu}
\orcid{https://orcid.org/0009-0007-2199-8518}
\affiliation{%
  \institution{University of Science and Technology of China}
  \institution{State Key Laboratory of Cognitive Intelligence}
  \city{Hefei}
  \country{China}}
\email{zihaowu@mail.ustc.edu.cn}

\author{Chaohu Liu}
\orcid{https://orcid.org/0009-0001-7588-4264}
\affiliation{%
  \institution{University of Science and Technology of China}
  \institution{State Key Laboratory of Cognitive Intelligence}
  \city{Hefei}
  \country{China}}
\email{liuchaohu@mail.ustc.edu.cn}

\author{Linli Xu}
\authornote{Corresponding author}
\orcid{https://orcid.org/0000-0003-0227-3793}
\affiliation{%
  \institution{University of Science and Technology of China}
  \institution{State Key Laboratory of Cognitive Intelligence}
  \city{Hefei}
  \country{China}}
\email{linlixu@ustc.edu.cn}

\renewcommand{\shortauthors}{YongXiang Hua,Haoyu Cao,Zhou Tao,Bocheng Li,Zihao Wu,Chaohu Liu and Linli Xu}

\begin{abstract}
Sparse Mixture of Experts (sMoE) has become a pivotal approach for scaling large vision-language models,
offering substantial capacity while maintaining computational efficiency through dynamic, sparse activation of experts.
However, existing routing mechanisms, typically based on similarity scoring, struggle to effectively capture the underlying input structure. 
This limitation leads to a trade-off between expert specialization and balanced computation, hindering both scalability and performance. 
We propose Input Domain Aware MoE, a novel routing framework that leverages a probabilistic mixture model to better partition the input space. 
By modeling routing probabilities as a mixture of distributions, our method enables experts to develop clear specialization boundaries while achieving balanced utilization. 
Unlike conventional approaches, our routing mechanism is trained independently of task-specific objectives, allowing for stable optimization and decisive expert assignments.
Empirical results on vision-language tasks demonstrate that our method consistently outperforms existing sMoE approaches, achieving higher task performance and improved expert utilization balance.
\end{abstract}




\maketitle

\section{Introduction}
\label{sec:intro}
Vision-language models (VLMs) have demonstrated remarkable capabilities, but their immense computational demands pose significant scaling challenges. 
To address this, sparsely-gated Mixture-of-Experts (sMoE) was introduced in Large Language Models (LLMs) \cite{lepikhin2020gshard} and has since established itself as a dominant solution, 
finding widespread applications across models of varying scales and domains\cite{dai2024deepseekmoe,zhang2024mm1,yu2023mmoe,wu2024omni,gao2024enhanced,awitm,hrvda}.
At its core, sMoE employs conditional computation where each input token activates only a subset of specialized experts through a learned routing function, 
effectively reducing computational overhead while maintaining model capacity. 
However, the effectiveness of this approach ultimately hinges on the routing mechanism that assigns inputs to experts.

\begin{figure}[t]
    \centering
    \begin{subfigure}{0.49\linewidth}
        \includegraphics[width=\linewidth]{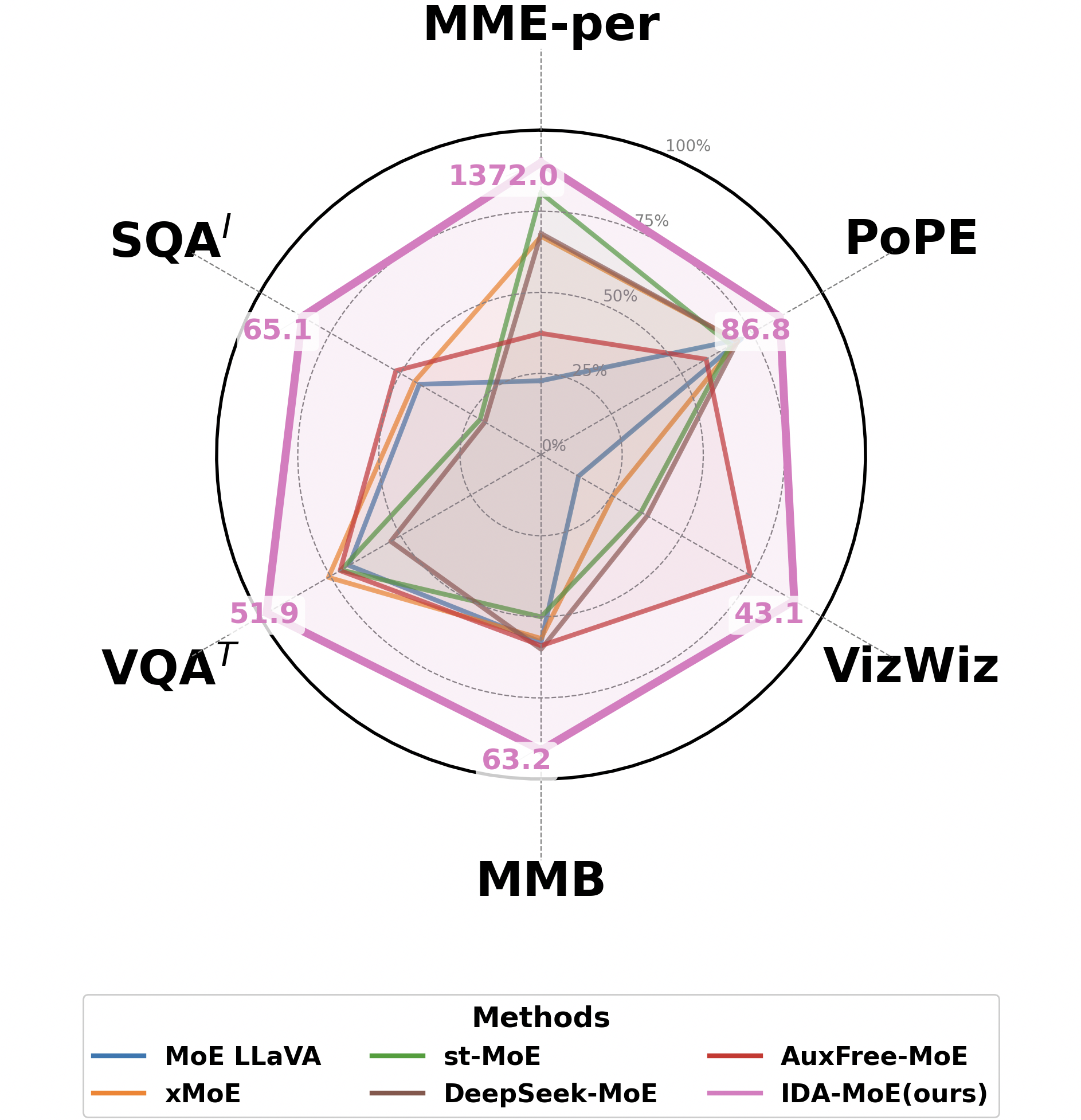}
        \caption{Comparative analysis of IDA-MoE against various routing mechanisms implemented in LLaVA MoE}
        \label{fig:ablation_radar}
    \end{subfigure}
    \hfill
    \begin{subfigure}{0.49\linewidth}
        \includegraphics[width=\linewidth]{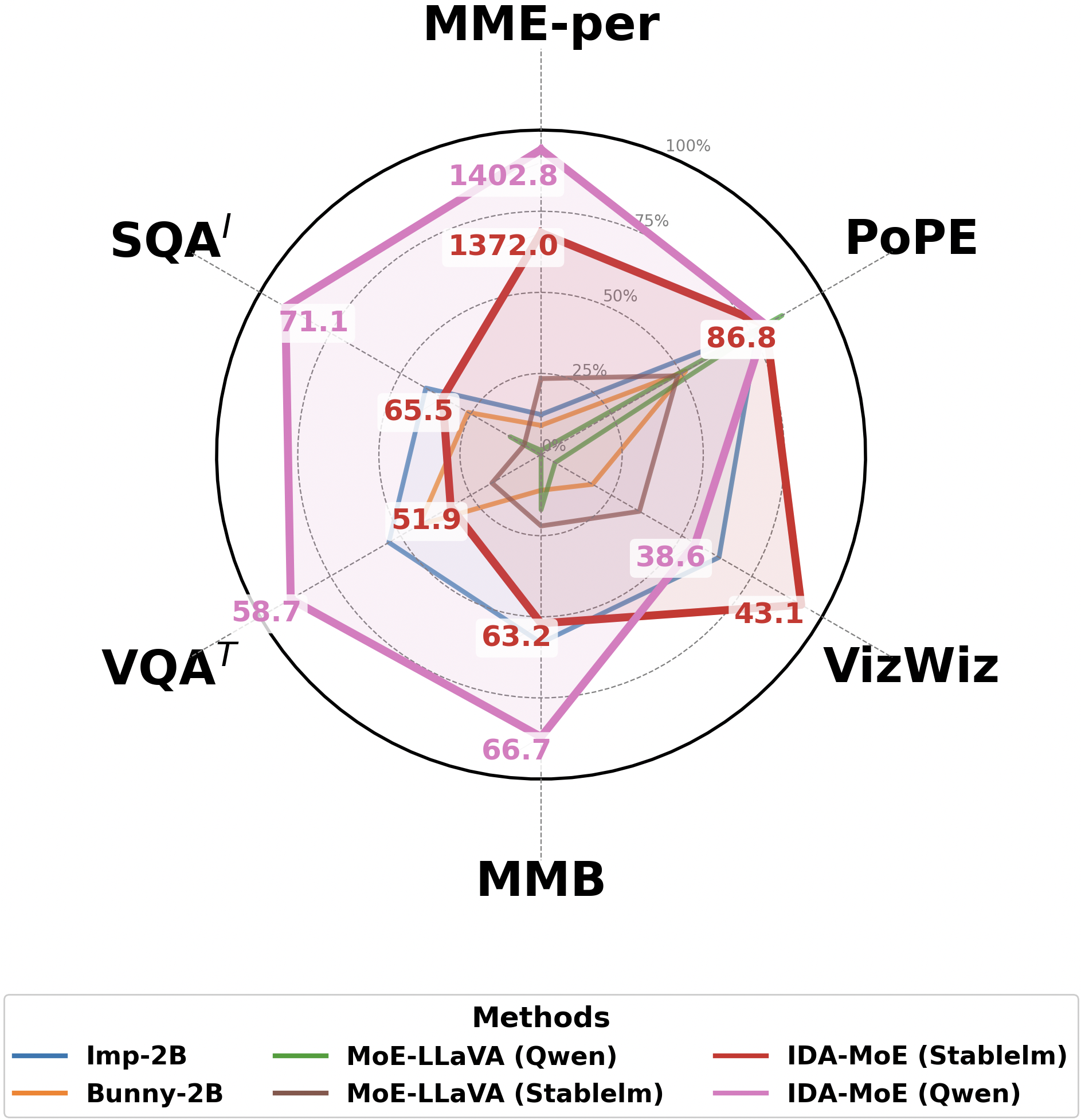}
        \caption{Performance evaluation of IDA-MoE versus sota open-source models of comparable parameter size (2B)}
        \label{fig:sota_radar}
    \end{subfigure}
    \caption{Benchmark performance of IDA-MoE (ours) across visual question answering and multimodal reasoning tasks. The left panel (\cref{fig:ablation_radar}) illustrates the effectiveness of different MoE routing implementations, while the right panel (\cref{fig:sota_radar}) demonstrates comparative performance against models of similar computational scale.}
    \label{fig:radar}
\end{figure}

Various routing methods have been proposed. Among them, 
the predominant approaches \cite{lepikhin2020gshard,fedus2022switch,dai2024deepseekmoe, liu2024deepseek,liu2024grin} rely on similarity-based scoring, 
where routing probabilities are computed through softmax over token-expert alignment scores. 
This mechanism, trained end-to-end with the task objective, 
tends to develop a strong bias towards routing tokens to high-performing experts, 
creating a feedback loop that exacerbates load imbalance. 
To mitigate this, conventional approaches introduce auxiliary load balancing objectives\cite{shazeer2017,lepikhin2020gshard,zoph2022st} 
that push per-token routing distributions toward uniformity across experts. 
However, this creates a fundamental contradiction: 
while the core principle of sMoE emphasizes specialized experts through conditional computation,
load balancing forces indiscriminate token distribution, inevitably undermining expert specialization. 
This unresolved tension manifests in two critical problems: 
during training, routing instability leads to frequent changes in token-expert assignments, severely impacting sample efficiency \cite{dai2022stablemoe, zoph2022st};
during inference, ambiguous routing decisions result in suboptimal expert selection, compromising model performance and robustness to distribution shifts \cite{wu2024gw,wang2024auxiliary}.

The root cause lies in coupling routing decisions with task optimization,
which makes it contradictory to achieve balanced token allocation while preserving sharp, decisive routing.
Specifically, when experts are evaluated solely on their contribution to the final loss,
the system naturally concentrates computation on a small subset of versatile experts rather than developing complementary specializations(analysed in \cref{subsec:imbalanced}).
This suggests a fundamental need to decouple routing from task-specific learning signals and instead ground it in the inherent structure of the input distribution.

To this end, we propose Input Domain Aware MoE (IDA-MoE), 
which explicitly models the input space through a Gaussian Mixture Model (GMM) to create natural, data-driven routing boundaries.
This probabilistic framework partitions inputs based on their distribution characteristics rather than loss contributions,
where routing decisions are guided by the posterior probabilities rather than learned similarity scores,
enabling experts to develop clear specializations while maintaining balanced utilization.
By associating each expert with multiple mixture components (i.e., regions of the input distribution) rather than learning routing directly from task objectives,
our approach creates a flexible mapping between input domains and expert specializations. 
This decoupling allows independent optimization of expert workload distribution (controlled by mixture priors $P(k)$) 
while maintaining sharp, distribution-aware token routing - achieving load balance as an emergent property rather than an enforced constraint.
Our contributions in this work are threefold. 

\begin{itemize}
    \item First, we present a systematic analysis of existing MoE routing mechanisms, revealing the fundamental dilemma: the coupling between routing and task objectives inherently drives load imbalance, while conventional mitigation strategies often compromise training stability and prevent effective expert specialization.
    \item Second, we introduce Input Domain Aware MoE (IDA-MoE), a novel architecture that explicitly models input distribution through probabilistic mixture modeling, effectively decoupling routing decisions from task optimization while maintaining precise expert specialization.
    \item Third, through comprehensive experiments across multiple vision-language tasks, we demonstrate that IDA-MoE achieves superior performance compared to traditional routing approaches, exhibiting improved training stability, better expert utilization, and enhanced performance on VQA benchmarks.
\end{itemize}

\section{Related Works}
\label{sec:RelatedWork}

\paragraph{Large Vision-Language Models.}
The emergence of powerful Large Language Models (LLMs) has enabled a new paradigm for multimodal learning,
where frozen LLMs act as reasoning engines for vision-language tasks. 
Early approaches like BLIP-2\cite{li2023blip} and LLaVA\cite{liu2023visual, liu2024improved} proposed lightweight adapters (e.g., Q-Former, MLP projectors)
to map visual features from pretrained encoders (e.g., CLIP\cite{radford2021learning}) into the LLM's token space.
This adapter-based paradigm has achieved substantial performance gains across diverse vision-language tasks.
Recent works \cite{young2024yi,fu2024vita,fu2025vita,shen2025long, chen2024expanding, bai2025qwen2, chen2024internvl, grattafiori2024llama} continue this paradigm by aggressively scaling both model parameters and training data.
While this leads to improved capabilities, it also makes training and inference prohibitively expensive.
To address these computational challenges, researchers have explored sparse Mixture of Experts (sMoE) architectures,
which are adopted both in pre-training \cite{bao2022vlmo, bai2025qwen2, wu2024deepseek} and via efficient upcycling of dense models \cite{komatsuzaki2022sparse, lin2024moe, li2024cumo, li2025uni, wudynamic, xu2024leveraging, shu2024llava} in multimodal context. 
Despite these advances, sMoE architectures face significant load balancing problems. 
This challenge is particularly pronounced in multimodal contexts, as inputs from the same modality tend to activate identical expert pathways, further exacerbating the load imbalance problem.

\paragraph{Mixture-of-Experts,}
particularly Sparse Mixture-of-Experts, was introduced to large language models by Gshard\cite{lepikhin2020gshard} and refined in subsequent works \cite{fedus2022switch, jiang2024mixtral,dai2024deepseekmoe,liu2024deepseek, wei2024skywork, xue2024openmoe}. 
This architecture enables capacity scaling while preserving computational efficiency through selective expert activation.
Since its introduction, sparse Mixture-of-Experts models have faced persistent challenges with the load imbalance problem.
The predominant approaches rely on \textbf{Auxiliary Losses}, which penalize imbalanced expert utilization based on token allocation fractions\cite{lepikhin2020gshard,fedus2022switch, liu2024deepseek}, Coefficient of Variance of token allocation\cite{shazeer2017} or routing entropy\cite{shen2023moduleformer}.
However, multiple studies \cite{dai2022stablemoe,wu2024gw,wang2024auxiliary} demonstrate that auxiliary losses can significantly impair training stability and model performance.
st-MoE\cite{zoph2022st} introduced Router-z Loss to stabilize training.
Recent approaches such as Auxiliary-free-MoE\cite{wang2024auxiliary} and Deepseek-v3\cite{liu2024deepseekv3} attempt to mitigate the side effects of load balancing loss through introducing prior distribution correction.
Another line of work achieves \textbf{Runtime Load Balancing} via token dropping \cite{fedus2022switch} or sophisticated system-level approaches that involve dynamic resource allocation or predictive scheduling\cite{nie2023flexmoe, wang2024pro, pmlr-v235-kim24w, fastermoe}.
Nevertheless, these methods have yet to achieve an optimal balance between computational efficiency and model performance in multimodal settings.
In contrast to previous approaches, our work addresses the fundamental tension between expert specialization and computational balance by introducing a novel routing mechanism that enables effective specialization while maintaining efficient resource utilization.

\section{The Specialization-Balance Dilemma}
\label{sec:background}
In this section, we delve into a fundamental challenge of standard Sparse Mixture-of-Experts(sMoE) architectures:
the inherent tension between achieving expert specialization and maintaining balanced computation.
We first provide the necessary background on the standard sMoE architecture and token-choice gating mechanism (\cref{subsec:sMoE}).
Then, we analyze how coupling routing decisions directly with task optimization promotes expert specialization but results in inevitable load imbalance (\cref{subsec:imbalanced}).
Finally, we examine how conventional auxiliary load balancing loss compromises specialization and stability, thus revealing the core dilemma (\cref{subsec:Specialization}).


\subsection{Sparse Mixture of Experts}
\label{subsec:sMoE}
In modern transformer architectures, Sparse Mixture of Experts (sMoE) represents a powerful approach to scaling model capacity while maintaining computational efficiency. sMoE typically replaces the standard feedforward network (FFN) with a set of expert networks, where each input token is dynamically routed to a small subset of these experts based on its content.

Consider a standard $L$-layer Transformer processing an input sequence $X \in \mathbb{R}^{T \times d}$ with $T$ tokens and embedding dimension $d$. At each layer $l$, let $\mathbf{h}^l_t$ represent the hidden state for token $t$, and $\mathbf{u}^l_t$ represent the intermediate representation after self-attention:
\begin{align}
   \mathbf{u}^l_{t} &= \text{SelfAttn}(\mathbf{h}^{l-1}) + \mathbf{h}^{l-1}_{t} \\
   \mathbf{h}^l_t &= \text{FFN}(\mathbf{u}^l_t) + \mathbf{u}^l_t
\end{align}
In sMoE-based transformers, the FFN module is replaced with a dynamic mixture of expert networks:
\begin{align}
   \mathbf{h}_t^l &= \sum_{i=1}^k \mathcal{G}(\mathbf{u}_t^{l}, \mathbf{e}_i) \cdot \text{FFN}_i(\mathbf{u}_t^{l}) + \mathbf{u}_t^{l}
\end{align}
where $\mathcal{G}(\cdot, \mathbf{e}_i)$ computes routing scores measuring token-expert compatibility, with $\mathbf{e}_i$ representing routing parameters of expert $i$, and $\text{FFN}_i(\cdot)$ represents the $i$-th expert network.
For simplicity, we note $MoE(\textbf{u}_t^{l}) = \sum_{i=1}^k \mathcal{G}(\mathbf{u}_t^{l}, \mathbf{e}_i) \cdot \text{FFN}_i(\mathbf{u}_t^{l})$.

The choice of gating function is crucial for dynamic expert selection.
The predominant approach in large language and vision-language models is Token-Choice Gating, 
where each token independently selects its preferred top-$k$ experts based on compatibility scores.
While alternative routing strategies 
exist, they present various trade-offs. For instance, 
Expert-Choice Gating \cite{zhou2022mixture,zhou2023brainformers} 
enforces uniform expert workload but risks uneven token coverage and violates the  
causal constraints of language modeling~\cite{wang2024auxiliary},
while Non-trainable Gating \cite{roller2021hash,zuo2021taming,ren2023pangu} offers balanced computation via static or randomized routing at the cost of adaptability.
Considering these trade-offs, our work focuses on Token-Choice Gating, which computes routing weights as:
\begin{equation}
   \mathbf{p}_t = \text{Softmax}(\mathbf{u}_t^{l-1} \cdot [\mathbf{e}_1, \ldots, \mathbf{e}_K])
\end{equation}
\begin{equation}
   \mathcal{G}(\mathbf{u}_t^{l-1}, \mathbf{e}_i) = \delta_{t,i} \cdot \mathbf{p}_{t,i}
\end{equation}
where $\mathbf{p}_t$ represents the routing distribution for token $t$, with each element $\mathbf{p}_{t,i}$ measuring the token-expert compatibility. 
$\delta_{t,i}$ is a binary indicator, which equals 1 
only when expert $i$ is among the top-$k$ highest-probability experts for token $t$. 

\subsection{Load Imbalance}
\label{subsec:imbalanced}

Load imbalance is a persistent and critical challenge in sMoE models. 
Numerous studies \cite{shazeer2017, lepikhin2020gshard,fedus2022switch, zhou2022mixture,dai2024deepseekmoe,wang2024auxiliary} observe that computation is often heavily skewed in practice, with a small subset of experts processing a disproportionate fraction of tokens while others remain largely inactive.
This disparity significantly compromises parameter efficiency, computation utilization, and the overall model scaling potential.
While previous works have 
attempted to mitigate this issue through auxiliary load balancing losses (discussed in \cref{subsec:Specialization}),
we demonstrate that load imbalance is not merely an implementation side-effect,
but an inherent consequence of the underlying sMoE training dynamics.
By analyzing how the task loss $\mathcal{L}$ influences the expert parameters $\textbf{e}_i$ , which in turn determine routing probabilities, we can reveal the fundamental mechanisms that drive this imbalance. Specifically, the
gradient of the task loss with respect to $\textbf{e}_i$ is given by:
\begin{equation}
    \label{equ:derivative}
    \nabla_{\mathbf{e}_i} \mathcal{L} = \sum_{t=1}^T \delta_{t,i} \mathbf{p}_{t,i} \cdot (\mu_{t,i} - \bar{\mu}_t) \cdot \mathbf{u}_t^{l-1}
\end{equation}
\begin{equation}
    \mu_{t,i} = \frac{\partial \mathcal{L}}{\partial \text{MoE}(\mathbf{u}_t^{l-1})} \cdot \text{FFN}_i(\mathbf{u}_t^{l-1})
\end{equation}
\begin{equation}
    \bar{\mu}_t = \frac{\partial \mathcal{L}}{\partial \text{MoE}(\mathbf{u}_t^{l-1})} \cdot \text{MoE}(\mathbf{u}_t^{l-1}) 
\end{equation}
where $\mu_{t,i}$ quantifies how effectively expert $i$ processes token $t$ through task-specific loss gradients, while $\bar{\mu}_t$ represents the average effectiveness of the activated experts for that token.
This gradient dictates updates to the routing parameters $\textbf{e}_i$ , thereby shaping future routing probabilities $\textbf{p}_{t',i}$ for similar tokens $t'$.

\begin{figure}[t]
    \vspace{-2mm}
    \centering
    \begin{subfigure}{0.48\linewidth}
        \includegraphics[width=\linewidth]{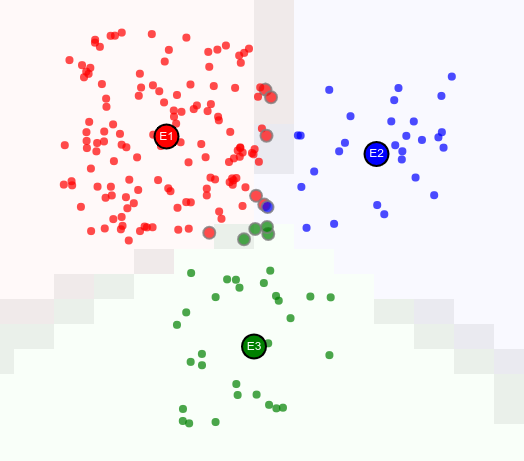}
        \caption{Token-Expert Allocation before load-balancing}
        \label{fig:before_lb}
    \end{subfigure}
    \hfill
    \begin{subfigure}{0.48\linewidth}
        \includegraphics[width=\linewidth]{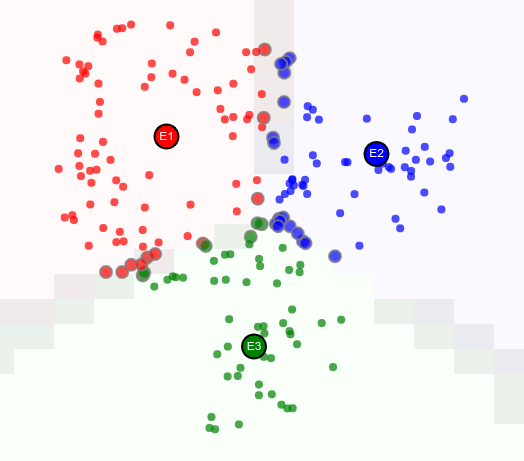}
        \caption{Token-Expert Allocation after load-balancing}
        \label{fig:after_lb}
    \end{subfigure}
    \caption{Illustration of load balancing loss effects. Experts colored circles and their corresponding token assignments colored dots in matching colors. 
    The transition from \cref{fig:before_lb} to \cref{fig:after_lb} simulates the effect of load 
    balancing loss. Token embeddings are pushed toward and accumulated around decision boundaries.}
    \label{fig:lb_illu}
    \vspace{-4mm}
\end{figure}

Careful examination of this
gradient reveals several interconnected factors that systematically drive expert 
imbalance. 1) \textbf{Relative Performance Steering}: Updates are steered by the relative performance metric $(\mu_{t,i} - \bar{\mu}_t)$, and $\textbf{e}_i$ is
preferentially updated towards alignment with tokens where expert $i$ outperforms its peers in loss reduction.
This leads to expert specialization but also concentrates tokens around already-competent experts. 
2) \textbf{Probability Amplification}: The expert concentration is amplified by the probability amplification factor $\textbf{p}_{t,i}$ in \cref{equ:derivative}, 
as experts already deemed more suitable for certain tokens receive proportionally stronger gradient updates,  
creating a multiplicative effect that rapidly amplifies small initial advantages. 
3) \textbf{Hard Gating Exclusion}: The Top-K selection mechanism exacerbates the imbalance by completely blocking updates for non-selected experts.
This prevents adaptation for less-utilized experts and can lead to permanently underutilized or ``dead'' experts, 
especially during distribution shifts in training. 

In conjunction, the mechanisms of performance biased steering, probability amplified updates, and hard gating interact to create a powerful positive feedback loop that systematically reinforces initial imbalances across the expert pool.
Experts gaining an early lead attract more tokens and stronger updates, solidifying their dominance, 
while less favored experts become progressively starved of both data and meaningful gradients,
leading to severe and potentially irreversible load imbalance.
This ``rich-get-richer'' dynamic demonstrates that load imbalance is fundamentally intertwined with optimizing routing directly via task performance feedback.

\subsection{The Specialization-Balance Dilemma}
\label{subsec:Specialization}

To counteract the inherent load imbalance described previously, the standard practice in sMoE models involves incorporating an auxiliary load balancing loss \cite{lepikhin2020gshard,fedus2022switch,jiang2024mixtral,lin2024moe,liu2024deepseek}.
One of the most prominent form is:
\begin{equation}
    \label{load_balancing_loss_function}
    \mathcal{L}_{\text {Balance }}  =\alpha \sum_{i=1}^{N} f_i \textbf{p}_i
\end{equation}
\begin{align}
    \label{load_balancing_loss_function_exp}
    f_i  &=\frac{N}{K T} \sum_{t=1}^T \delta_{t ,i}\\
    \textbf{p}_i  &=\frac{1}{T} \sum_{t=1}^T \textbf{p}_{t, i}
\end{align}
where $N$ is the total number of experts, $K$ is the number of experts selected for each token, $f_i$ represents the fraction of tokens routed to expert $i$, and $\alpha$ is a hyper-parameter controlling the strength of the auxiliary loss.

This balancing loss introduces a fundamental \textbf{specialization-balance dilemma}: while effective expert specialization requires the router to make decisive assignments through sharply peaked routing probabilities $\textbf{p}_{t,i}$ (directing each token confidently to its most appropriate experts), the auxiliary loss actively counteracts this tendency by penalizing concentrated allocations to achieve more balanced token distribution across the expert pool.

Specifically, auxiliary loss seeks to balance the workload $f_i$ across experts primarily by penalizing high routing probabilities $\textbf{p}_{t,i}$.
As illustrated in \cref{fig:lb_illu}, 
this loss redistributes tokens from overloaded experts to underutilized ones by pushing them across decision boundaries. 
However, this redistribution creates an unintended side effect: tokens tend 
to accumulate near routing decision boundaries 
(\cref{fig:after_lb}), leading to ambiguous expert assignments and suboptimal performance during both training and inference.

\begin{figure}[t]
    \centering
    \setlength{\abovecaptionskip}{0.cm}
    \begin{subfigure}[b]{0.9\columnwidth}
        \centering
        \includegraphics[width=\linewidth]{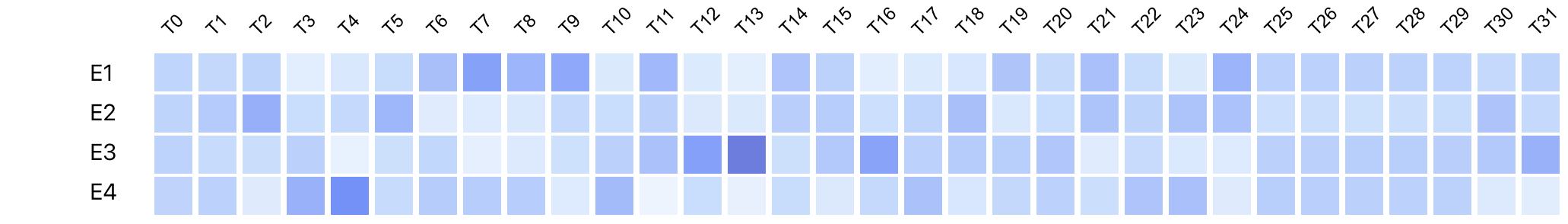}
        \caption{Routing Probability of MoE LLaVA}
        \label{fig:routing_aux}
    \end{subfigure}
    \vspace{2mm}
    \begin{subfigure}[b]{0.9\columnwidth}
        \centering
        \includegraphics[width=\linewidth]{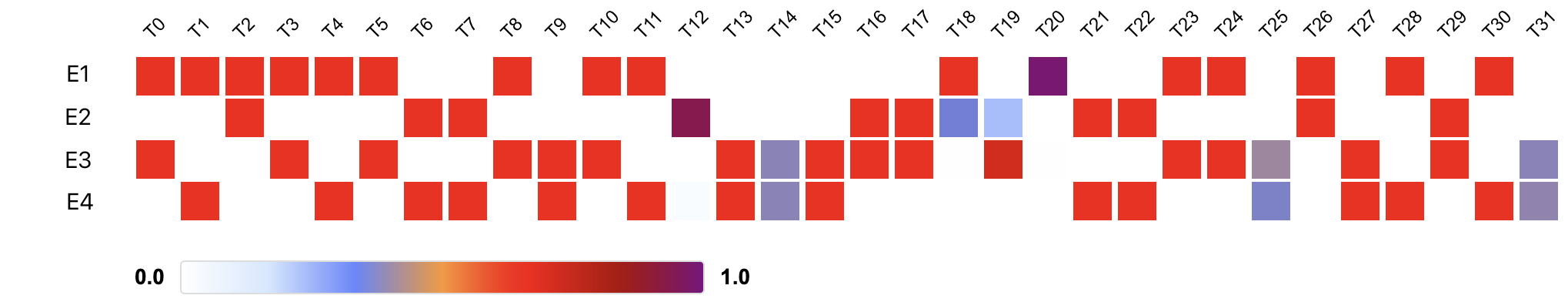}
        \caption{Routing Probability of IDA MoE}
        \label{fig:routing_ida}
    \end{subfigure}
    \caption{Routing probability heatmaps of the last MoE layer for the same input sequence. The x-axis represents token indices, the y-axis represents expert indices, and the color intensity indicates the probability of assigning the $i$-th token to the $j$-th expert. Note how IDA-MoE (b) exhibits significantly more decisive routing patterns with clearer token-expert affinities compared to the conventional approach (a). }
    \label{fig:routing_probability}
    \vspace{-3mm}
\end{figure}

\begin{figure*}[t]
    \centering
    \includegraphics[width=\linewidth]{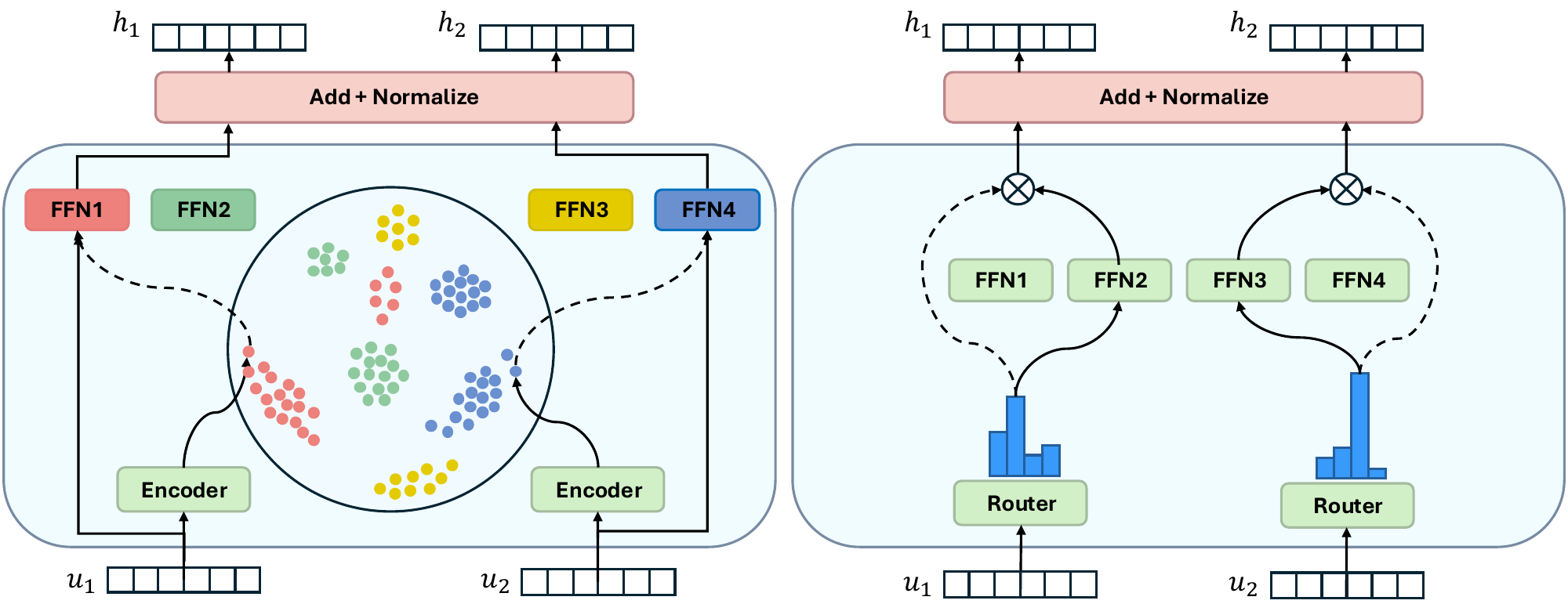}
    \caption{Comparison between IDA-MoE and traditional MoE routing. (Left) IDA-MoE: Input token representations ($\textbf{u}_1,\textbf{u}_2$ ) are projected into a lower-dimensional routing space where their distribution is modeled into distinct clusters. Tokens are routed to specific experts based on their cluster index, effectively decoupling routing decisions from 
    task-specific optimization. (Right) Traditional MoE: A learned router module directly calculates token-expert affinities (classification-style) to determine expert assignments, typically trained end-to-end with the task objective. }
    \label{fig:Stucture}
    \vspace{-4mm}
\end{figure*}

Specifically, during 
training, tokens
near boundaries may frequently switch assigned experts due to the competing pressures of task optimization and the load balancing loss,
hindering both training stability and expert specialization.
During inference, 
as illustrated in \cref{fig:routing_probability}, auxiliary losses typically 
produce high-entropy routing distributions, 
where token-expert assignment probabilities appear less distinct compared to an ideal sharp routing (\cref{fig:routing_ida}).
Such ambiguous routing creates inherent robustness issues. 
As prior research~\cite{wu2024gw,wang2024auxiliary} suggest, models with less decisive routing are more susceptible to performance degradation under input noise or distribution shifts, 
as small perturbations can easily alter routing decisions and consequently change model behavior.

Therefore, while conventional auxiliary losses successfully enforce load balance, 
they do so by fundamentally compromising the model's ability to specialize effectively.
The resulting trade-off 
negatively impacts both training stability and inference robustness.
This highlights the critical need for alternative routing strategies that can achieve balanced computation by directly addressing the input space structure, 
thereby preserving the sharpness and integrity of specialized expert routing.

\section{Method}

We propose Input Domain-Aware Mixture of Experts (IDA-MoE), a novel approach designed to achieve effective expert specialization and natural load balancing by decoupling routing decisions from task optimization and instead grounding them in the inherent structure of 
the input token distribution.
Our method consists of three key components: 
(1) A Gaussian Mixture Model that captures the underlying distribution of input tokens in a low-dimensional space,
(2) 
A component-based expert routing mechanism that leverages this probabilistic distribution model to make more informed and stable routing decisions, 
and (3) A 
component reactivation strategy that accelerates model convergence by addressing the uneven learning dynamics. 

\subsection{Decoupled Input Distribution Modeling}
\label{subsec:GMM_Modeling}

To efficiently model the input distribution while avoiding 
the curse of dimensionality, we first project the token representations $u^{l-1}$ into a lower-dimensional latent space using a simple autoencoder network:
\begin{equation}
    \textbf{z}_t = \text{Encoder}(\text{sg}(\textbf{u}^{l-1}))
    \label{eq:encoder}
\end{equation}
where sg() stand for stop gradient. This encoder is trained concurrently with the main model using a reconstruction loss to preserve the underlying input structure:
\begin{equation}
    \mathcal{L}_{\text{AE}} = \|\text{sg}(\textbf{u}^{l-1}) - \text{Decoder}(\textbf{z}_t)\|^2
    \label{eq:recon_loss}
\end{equation}

Within this compressed routing space, we model the input data distribution using a Gaussian Mixture Model (GMM). 
To enable experts to capture the intricacy of the underlying input structure, we 
associate the input domain of each expert with $M$ fine-grained GMM components rather than a single Gaussian distribution.
Thus, the overall distribution of the routing space can be formulated as:
\begin{equation}
    p(\textbf{z}_t) = \sum_{i=1}^{N} \sum_{m=1}^{M} \boldsymbol{\pi}_{i,m} \mathcal{N}(\textbf{z}_t | \boldsymbol{\mu}_{i,m}, \boldsymbol{\Sigma}_{i,m})
    \label{eq:expert_gmm}
\end{equation}
where $\boldsymbol{\pi}_{i,m}$ denotes the mixing coefficient, and $\boldsymbol{\mu}_{i,m}$, $\boldsymbol{\Sigma}_{i,m}$ are the mean and covariance of the $m$-th component for the $i$-th expert.

The GMM is trained jointly with the backbone model 
by minimizing the 
negative log-likelihood 
of the observed token representations under the mixture distribution:
\begin{equation}
    \mathcal{L}_{\text{GMM}} = - \sum_{t=1}^{T} \log \left( \sum_{i=1}^{N} \sum_{m=1}^{M} \boldsymbol{\pi}_{i,m} \mathcal{N}(\text{sg}(\textbf{z}_t) | \boldsymbol{\mu}_{i,m}, \boldsymbol{\Sigma}_{i,m}) \right)
    \label{eq:nll_loss}
\end{equation}
\vspace{-2mm}
\subsection{Component-Based Expert Routing}

Standard MoE layers typically route each token to $k \geq 1$ experts. 
To implement top-$k$ routing, IDA-MoE 
employs $k$ independent sets of expert-associated GMM parameters as described in \cref{subsec:GMM_Modeling}. 
Each GMM set $j \in \{1, \dots, k\}$ contains parameter triplets $(\boldsymbol{\pi}_{j,i,m}, \boldsymbol{\mu}_{j,i,m}, \boldsymbol{\Sigma}_{j,i,m})$ and is trained using 
an independent NLL loss $\mathcal{L}_{\text{GMM}_j}$. 
This allows IDA-MoE to make $k$ distinct expert selections for each token based on potentially different distributional perspectives captured by each GMM set.
This implementation 
introduces minimal overhead due to the low dimensionality of $\textbf{z}_t$.
The routing process for a token $\textbf{u}_t$ with latent representation $\textbf{z}_t$ proceeds as follows:

\noindent\textbf{Calculate Posteriors:} For each selection rank $j$, we use the corresponding $j$-th GMM parameter set to compute the posterior probability $P_j(i,m|\textbf{z}_t)$. This posterior quantifies 
the likelihood that the latent token $\textbf{z}_t$ belongs to component $m$ associated with expert $i$:
\begin{equation}
    P_j(i, m | \textbf{z}_t) = \frac{\boldsymbol{\pi}_{j,i,m} \mathcal{N}(\textbf{z}_t | \boldsymbol{\mu}_{j,i,m}, \boldsymbol{\Sigma}_{j,i,m})}{\sum_{i'=1}^{N} \sum_{m'=1}^{M} \boldsymbol{\pi}_{j,i',m'} \mathcal{N}(\textbf{z}_t | \boldsymbol{\mu}_{j,i',m'}, \boldsymbol{\Sigma}_{j,i',m'})}
    \label{eq:posterior} 
\end{equation}
\textbf{Select Top Experts:} For each rank $j$, we determine the expert $i^*_{t,j}$ whose components maximize the posterior probability for $\textbf{z}_t$. 
\begin{align}
    P^*_j(i, \textbf{z}_t) &= \max_{m \in \{1,\dots,M\}} P_j(i, m | \textbf{z}_t) \quad \forall i \in \{1,\dots,N\} \label{eq:max_posterior_per_expert}\\
    i^*_{t,j} &= \arg\max_{i \in \{1,\dots,N\}} P^*_j(i, \textbf{z}_t) \label{eq:select_expert_index}
\end{align}
This yields a set of $k$ selected expert indices $\mathcal{I}_t = \{i^*_{t,1}, \dots, i^*_{t,k}\}$ and a corresponding vector of their maximum posterior scores $\textbf{P}^*_t = [P^*_1(i^*_{t,1}, \textbf{z}_t), \dots, P^*_k(i^*_{t,k}, \textbf{z}_t)]$.

\noindent\textbf{Calculate Gating Weights:} We compute the final gating weights by applying the softmax function across the vector of maximum posterior scores $\textbf{P}^*_t$ corresponding to the $k$ selected experts:
\begin{equation}
    \mathcal{G}_t = softmax(\textbf{P}^*_t)
    \label{eq:softmax_gates}
\end{equation}
\textbf{Final Output Calculation:}
The output of the IDA-MoE layer for token $\textbf{u}_t$ is the weighted sum of the outputs from the $k$ selected expert networks, using the calculated gating weights:
\begin{equation}
    \text{MoE}(\textbf{u}_t) = \sum_{j=1}^{k} \mathcal{G}_{t,j} \cdot \text{FFN}_{i^*_{t,j}}(\textbf{u}_t)
    \label{eq:final_moe_output_corrected}
\end{equation}
These gating weights are derived directly from the input distribution via the GMM posteriors, ensuring that routing remains decoupled from the final task optimization, breaking the problematic feedback loop described in \cref{subsec:imbalanced}.

\begin{table*}[!t]
    \setlength{\abovecaptionskip}{0.cm}
    \caption{
        Comparison with state-of-the-art MLLMs on the commonly-used multimodal benchmarks for MLLMs. 
        \textbf{Vis.} refers to the vision encoder and image resolution,
        \textbf{Act.} refers activated parameters during inference,
        The best result for model sizes around 2B is shown in bold, and the second-best result is underlined.
    }

    \label{tab:mainresult_hori}
    \begin{center}
        \setlength{\tabcolsep}{1.6mm}
        \begin{tabular}{l|ccc|cccccccc}
            \toprule
            \textbf{Method}                                         & \textbf{LLM}    & \textbf{Vis.} & \textbf{Act.} & \textbf{GQA}     & \textbf{PoPE}    & \textbf{VisWiz} & \textbf{SQA$^\text{I}$} & \textbf{TextVQA} & \textbf{MME}    & \textbf{MMB}  & \textbf{MMVET} \\
            \midrule
            \rowcolor{gray!20} {VILA-7B}\cite{lin2024vila}          & {LLaMA-7B}      & CLIP-336      & 7B            & {62.3}           & {-}              & {57.8}          & {68.2}                  & {64.4}           & {1533.0}        & {68.9}        & {-}            \\

            \rowcolor{gray!20} {InstructBLIP}\cite{liu2023visual}   & {Vicuna-13B}    & Vit-224      & 13B           & {49.5}           & {-}              & {33.4}          & {63.1}                  & {50.7}           & {1212.8}        & {-}           & {-}            \\

            \rowcolor{gray!20} {Qwen-VL-Chat}\cite{bai2023qwen}     & {Qwen-7B}       & Vit-448       & 7B            & {57.5}           & {-}              & {38.9}          & {68.2}                  & {61.5}           & {1487.5}        & {60.6}        & -              \\


            \rowcolor{gray!20} {LLaVA-1.5-7B}\cite{liu2023improved} & {Vicuna-1.5-7B} & CLIP-336      & 7B            & {62.0}           & {86.7}           & {50.0}          & {68.4}                  & {58.2}           & {1476.9}        & {61.5}        & 30.2           \\

            \midrule

            \rowcolor{gray!20} Bunny-3B\cite{he2024efficient}       & Phi-2-2.7B      & SigLIP-384    & 3.1B          & {62.5}           & {86.8}           & {43.8}          & {70.9}                  & {56.7}           & 1488.8          & 68.6          & -              \\

            \rowcolor{gray!20} LLaVA-Phi\cite{zhu2024llava}         & Phi-2-2.7B      & CLIP-336      & 2.8B          & {-}              & {85.0}           & {35.9}          & {68.4}                  & {48.6}           & {1335.1}        & {59.8}        & {28.9}         \\

            \rowcolor{gray!20} VILA-3B\cite{lin2024vila}            & LLaMA-2.7B      & SigLIP-384    & 2.7B          & {61.5}           & {-}              & {53.5}          & {69.0}                  & {60.4}           &      {-}        & 63.4          & -              \\

            \rowcolor{gray!20} Mini-Gemini\cite{li2024mini}      & Gemma-2B        & SigLIP-384    & 2.7B          & 60.7             & {-}              & {41.5}          & 63.1                    & {56.2} & {1341}          & 59.8          & {31.1}         \\

            \rowcolor{gray!20} MoE-LLaVA\cite{lin2024moe}           & Phi-2-2.7B      & CLIP-336      & 3.6B          & {61.4}           & {86.3}           & {43.9}          & {68.5}                  & {51.4}           & 1423.0          & 65.2          & 34.3           \\


            \midrule
            Imp-2B\cite{shao2024imp}                                & Qwen-1.5-1.8B   & SigLIP-384    & 1.9B          & \underline{61.9} & {86.7}           & \underline{39.6}          & \underline{66.1}        & \underline{54.5}           & 1304.8          & \underline{63.8}          & {-}            \\

            Bunny-2B\cite{he2024efficient}                          & Qwen-1.5-1.8B   & SigLIP-384    & 1.9B          & {59.6}           & {85.8}           & {34.2}          & {64.6}                  & {53.2}           & 1300.8          & 59.1          & {-}            \\



            MoE-LLaVA\cite{lin2024moe}                              & Qwen-1.8B       & CLIP-336      & 2.2B          & {61.5}           & \textbf{87.0}    & {32.6}          & {63.1}                  & {48.0}           & 1291.6          & 59.7          & {25.3}         \\

            MoE-LLaVA\cite{lin2024moe}                              & Stablelm-1.6B   & CLIP-336      & 2.0B          & {60.2}           & {85.7}           & {36.2}          & {62.6}                  & {50.1}           & {1318.1}        & 60.2          & {26.9}         \\

            \rowcolor{myblue} {IDA-MoE(ours)}                      & {Stablelm-1.6B} & CLIP-336      & 2.0B          & {58.2}          & \underline{86.8} & \textbf{43.1}  & {65.1}                  & {51.9}          & \underline{1372.0}        & {63.2}        & \underline{29.5}         \\

            \rowcolor{myblue} {IDA-MoE(ours)}                      & {Qwen2-1.5B}    & SigLIP-384    & 2.0B          & \textbf{62.1}    & \underline{86.8} & {38.6}         & \textbf{71.1}           & \textbf{58.7}   & \textbf{1402.8} & \textbf{66.7} & \textbf{33.4}  \\
            \bottomrule
        \end{tabular}
    \end{center}
    \vspace{-1mm}
\end{table*}


\subsection{Component Reactivation Strategy}
\label{sec:reactivation}
GMM training via minimizing the NLL loss (Eq.\eqref{eq:nll_loss}) 
can suffer from uneven convergence rates across mixture components.
Consider the gradient with respect to the mean $\boldsymbol{\mu}_{k}$ of a specific component $k = (j, i, m)$:
\begin{equation}
    \frac{\partial \mathcal{L}_{\text{GMM}}}{\partial \boldsymbol{\mu}_{k}} = - \sum_{t=1}^{T} \underbrace{\left( \frac{\boldsymbol{\pi}_{k} \mathcal{N}(\textbf{z}_t | \boldsymbol{\mu}_{k}, \boldsymbol{\Sigma}_{k})}{p(\textbf{z}_t)} \right)}_{\textbf{p}_{tk} \text{: Posterior}} \boldsymbol{\Sigma}_{k}^{-1} (\textbf{z}_t - \boldsymbol{\mu}_{k})
    \label{eq:gmm_gradient}
\end{equation}

As shown in Eq.\eqref{eq:gmm_gradient}, the gradient 
for component $k$ is weighted by its posterior probability given the data $\textbf{z}_t$. 
Components initialized in sparse regions of the feature space 
may consistently yield very low posterior probabilities ($\textbf{p}_{tk} \approx 0$),
leading to extremely slow convergence for those parameters, hindering the GMM's ability to fully capture the complete input distribution efficiently. 

To accelerate the convergence of these slowly updating components, we introduce a targeted reactivation strategy. 
We first identify potentially inactive components signaled by their low mixing coefficient $\boldsymbol{\pi}_{k}$, 
which often correlate with low average posterior probabilities, using a stochastic check:
\begin{equation}
    \text{is\_slow}(k) \sim \text{Bernoulli}(\text{ReLU}(1 - N \times M \cdot \boldsymbol{\pi}_{k}))
    \label{eq:is_slow_check}
\end{equation}
For the subset $S_j$ of components flagged as slow within each GMM set $j$, we apply a targeted \textbf{reactivation loss term}:
\begin{equation}
    \mathcal{L}_{\text{react}, j} = - \sum_{t=1}^{T} \log \left( \sum_{k \in S_j} \boldsymbol{\pi}_{k} \mathcal{N}(\text{sg}(\textbf{z}_t) | \boldsymbol{\mu}_{k}, \boldsymbol{\Sigma}_{k}) \right)
    \label{eq:react_loss}
\end{equation}
By calculating gradients based on Eq.\eqref{eq:react_loss}, we mathematically \textbf{renormalize the posterior probabilities within the subset $S_j$}. 
The new objective $\mathcal{L}_{\text{react}}$ provides substantial gradient updates to slow-updating components, 
pulling them towards relevant data points more quickly than $\mathcal{L}_{\text{GMM}}$ alone could, thus accelerating their convergence. 
Unlike conventional auxiliary losses that impose explicit balancing penalties and potentially disrupt expert specialization, 
this reactivation technique merely accelerates the natural convergence of the GMM towards a better fit of the input distribution. This approach preserves the specialized nature of expert routing while ensuring all components actively contribute to the model.

\textbf{Training Objectives.} The entire IDA-MoE model, including the backbone, autoencoder, and GMM parameters, is trained end-to-end by minimizing the composite loss function $\mathcal{L}_{total}$:
\begin{equation}
    \mathcal{L}_{total} = \mathcal{L}_{CE} + \alpha \cdot \mathcal{L}_{AE} + \beta \cdot \sum_{j=1}^k(\mathcal{L}_{GMM,j} + \mathcal{L}_{react,j})
\end{equation}
where $\mathcal{L}_{CE}$ is the cross-entropy loss for language modeling. 
The hyperparameters $\alpha$ and $\beta$ control the relative weight of the representation learning and distribution modeling objectives.
\vspace{-2mm}
\section{Experiments}
\label{sec:experiments}

\subsection{Experimental Setup}
\textbf{Model \& Training Setup.}
Our IDA-MoE model builds upon LLaVA architecture \cite{liu2023visual, liu2024improved}.
The MoE layers use top-2 routing, a latent dimensionality of 32, and 16 GMM components per expert with diagonal covariance for efficiency.
We set both $\alpha$ and $\beta$ to 0.01.
Training follows the three-stage protocol detailed in MoE-LLaVA \cite{lin2024moe}.
We bootstrap the dense model and GMMs in Stages 1 \& 2 using LLaVA-1.5-558k \cite{liu2023improved}, SViT \cite{zhao2023svit}, LVIS \cite{wang2023see}, LRV \cite{liu2023aligning}, and MIMIC-IT \cite{li2023mimic}.
In Stage 3, we finetune IDA-MoE follow the same data pipeline as LLaVA-mix-665k\cite{liu2023improved}. 
Further configuration and training specifics are detailed in the Supplementary Material Section 1 \& 2.

\textbf{Evaluation \& Baselines.}
We assess the performance across a comprehensive benchmark suite including MME \cite{liang2024survey}, MMB \cite{liu2024mmbench}, VizWiz \cite{gurari2018vizwiz}, GQA \cite{hudson2019gqa}, TextVQA \cite{singh2019towards}, MM-Vet \cite{yu2023mm}, ScienceQA \cite{lu2022learn}, and POPE \cite{li2023evaluating}.
Load balance is quantified using the mean Coefficient of Variation ($\mathrm{CV}_\mathrm{mean}$) of expert utilization (lower indicates better balance).
For fair comparison with state-of-the-art VLMs, 
we selected models based on comparable parameter scales and similar training data size and methodologies.
To further isolate the impact of our routing strategy,
we also compare against different routing methods using identical backbone model and training data. 
Results for the dense baseline and MoE-LLaVA \cite{lin2024moe} are reported from \cite{lin2024moe},
while other MoE implementations (xMoE \cite{chi2022representation}, st-MoE \cite{zoph2022st}, DeepSeek-MoE \cite{dai2024deepseekmoe}, AuxFree-MoE \cite{wang2024auxiliary}) are reproduced by us for fair comparison. 
Metric calculation details and reproduction specifics are provided in the Supplementary Material Section 3.

\begin{table*}
    \setlength{\abovecaptionskip}{0.1cm}
    \caption{
     Performance comparison of MoE approaches. All methods use Stablelm-1.6B and CLIP-336 as base models.
    }
    \label{tab:mainresult_verti}
    \begin{tabular}{l|cccccccc|c}
    \toprule
    \textbf{Method}                       & \textbf{MMVet} & \textbf{MME-per} & \textbf{PoPE} & \textbf{GQA}  & \textbf{SQA$^\text{I}$} & \textbf{TextVQA} & \textbf{VizWiz} & \textbf{MMB}  & $\mathbf{CV}_{mean}$ \\
    \midrule
    Dense                                 & 26.6           & 1337.3           & 85.6          & 60.6          & 62.7                    & 49.9            & 38.9            & 58.2          & -                \\
    MoE LLaVA\cite{lin2024moe}            & 26.9           & 1318.2           & 85.7          & 60.2          & 62.6                    & 50.1             & 36.2            & 60.2          & 0.128            \\
    xMoE\cite{chi2022representation}      & 28.5           & 1353.8           & 85.6          & {60.9}        & 62.7                    & 50.5            & 37.3            & 60.1          & 0.117            \\
    st-MoE\cite{zoph2022st}               & 29.3           & 1364.5           & 85.4          & {60.9}        & 61.3                    & 50.3            & 38.2            & 59.5          & 0.105            \\
    DeepSeek-MoE\cite{dai2024deepseekmoe} & 28.7           & 1354.5           & 85.6          & 60.7          & 61.2                    & 49.2            & 38.4            & 60.4          & 0.530            \\
    AuxFree-MoE\cite{wang2024auxiliary}   & 29.4           & 1329.9           & 84.7          & \textbf{61.1} & 63.1                    & 50.3            & 41.7            & 60.3          & 0.431            \\
    \rowcolor{myblue} IDA-MoE(ours)      & \textbf{29.5}  & \textbf{1372.0}  & \textbf{86.8} & 58.3         & \textbf{65.1}           & \textbf{51.9}   & \textbf{43.1}  & \textbf{63.2} & 0.143            \\
    \bottomrule
    \end{tabular}
    \vspace{-2mm}
\end{table*}

\subsection{Main Results}

\paragraph{Comparison with State-of-the-Art Models}
\cref{tab:mainresult_hori} presents a comprehensive comparison of IDA-MoE with recent state-of-the-art VLMs across multiple benchmarks.
Despite having significantly fewer activated parameters, our 2B parameter models achieve competitive performance with much larger dense models.
Our Qwen2-1.5B variant performs well against 7B models like LLaVA-1.5-7B and Qwen-VL-Chat, 
while outperforming several 3B models such as LLaVA-Phi ,Mini-Gemini and MoE-LLaVA-Phi on multiple benchmarks despite using only 2B activated parameters.
Within the 2B parameter category, IDA-MoE with Qwen2-1.5B clearly demonstrates state-of-the-art performance,
ranking first on six benchmarks and second on PoPE.
While our StableLM-1.6B variant also shows strong results, particularly on VisWiz benchmark.

\paragraph{Comparison Across MoE Implementations}
As shown in \cref{tab:mainresult_verti}, we compare different MoE methods on the same LLaVA backbone to isolate the impact of routing and balancing mechanisms.
Methods that enforce strict auxiliary losses, such as MoE-LLaVA, xMoE, and st-MoE, consistently produce low $\mathrm{CV}_\mathrm{mean}$ values, indicating excellent load balance.
However, MoE-LLaVA, which adopts standard auxiliary loss, shows only marginal performance gains over the dense baseline.
st-MoE introduces router-z loss to stabilize training, and xMoE employs dimensionality reduction combined with cosine-similarity-based routing to address representation collapse.
Although these enhancements allow st-MoE and xMoE to yield better results than the vanilla MoE setup,
their fundamental reliance on the auxiliary balancing objective still prevents them from fully exploiting the potential capacity benefits inherent in the MoE architecture.
In contrast, DeepSeek-MoE uses a significantly smaller load balancing coefficient (0.001 vs. the typical 0.01),
while AuxFree-MoE eliminates auxiliary losses entirely, relying instead on prior distribution correction for load balancing.
These methods relax balancing constraints to yield better task performance but suffer from substantial load imbalance (high $\mathrm{CV}_\mathrm{mean}$).
IDA-MoE stands out by achieving superior performance across most benchmarks while maintaining effective load balance ($\mathrm{CV}_\mathrm{mean}$: 0.1437) without relying on auxiliary losses.
This demonstrates the effectiveness of our distribution-driven, decoupled routing strategy in mitigating the specialization-balance dilemma.

\subsection{Ablation Study}


\begin{table}[h]
    \centering
    \caption{Ablation study on dimensionality reduction.}
    \vspace{-2mm}
    \label{tab:dim_ablation}
    \resizebox{\columnwidth}{!}{
        \begin{tabular}{cc|cccc|c}
            \toprule
            \textbf{Centers} & \textbf{Dim}  & \textbf{MME} &  \textbf{TextVQA} & \textbf{VizWiz} & \textbf{SQA$^\text{I}$} & $\mathbf{\mathrm{CV}}_\mathrm{mean}$ \\
            \midrule
            16 &  4  & 1340.7 & 49.95 & 36.77 & 61.73 & 0.1391 \\
            16 &  8  & 1363.4 & 50.53 & 36.84 & 62.22 & 0.1430 \\
            16 &  16 & 1365.4 & 50.40 & 38.40 & 62.82 & 0.1434 \\
            16 &  32 & \textbf{1372} & \textbf{51.86} & \textbf{43.11} & \textbf{65.05} & 0.1437  \\
            16 &  64 & 1364.1 & 50.29 & 40.61 & 63.61 & 0.1492 \\
            \bottomrule
        \end{tabular}
    }
    \vspace{-3mm}
\end{table}

\paragraph{Impact of routing dimensionality}
Table \ref{tab:dim_ablation} illustrates impact of the routing space dimensionality on model performance and load balance.
First, we observe that routing requires sufficiently high dimensionality to maintain discriminative power. 
While lower dimensions (4-8) produce slightly more balanced expert utilization, 
they significantly compromise performance across all benchmarks. 
This suggests that overly simplified representations lack the capacity to effectively distinguish between different input patterns, 
which make the token allocation more even but also leading to suboptimal expert assignment.
Second, we find that performance gains saturate and eventually decline as dimensionality increases beyond an optimal point.
This aligns with the well-known curse of dimensionality in mixture models,
where high-dimensional spaces become increasingly sparse, making density estimation less reliable for routing decisions.
This results identify 32 dimensions as the sweet spot that maximizes model performance while maintaining effective load distribution across experts and minimum overhead.

\begin{table}[h]
    \centering
    \caption{Ablation study on the number of GMM centers.}
    \vspace{-2mm}
    \label{tab:center_ablation}
    \resizebox{\columnwidth}{!}{
        \begin{tabular}{cc|cccc|c}
            \toprule
            \textbf{Centers} & \textbf{Dim} & \textbf{MME} & \textbf{TextVQA} & \textbf{VizWiz} & \textbf{SQA$^\text{I}$} & $\mathbf{\mathrm{CV}}_\mathrm{mean}$ \\
            \midrule
            1 & 32 & 1329.83 & 49.69 & 38.56 & 63.01 & 0.3218 \\
            4 & 32 & 1344.62 & 49.83 & 39.91 & 63.41 & 0.1974 \\
            16 & 32 & \textbf{1372} & \textbf{51.86} & \textbf{43.11} & \textbf{65.05} & 0.1437 \\
            32 & 32 &  1340.31 & 49.43 & 36.18 & 63.51 & 0.1364 \\
            \bottomrule
        \end{tabular}
    }
    \vspace{-2mm}
\end{table}
\paragraph{Impact of GMM components per expert}
Table \ref{tab:center_ablation} presents the 
impact of varying the number of GMM centers for the IDA-MoE model.
The 16-center 
configuration achieves substantial improvements over the single-center baseline: 
+42.17 points on MME, +2.17 points on TextVQA, +4.55 points on VizWiz, and +2.04 points on SQA.
These consistent gains across all benchmarks validate the effectiveness of multi-center approach for modeling token distributions. 
However, further increasing 
the number of centers to 32 leads to performance degradation across all tasks, suggesting a potential overfitting effect where additional components may 
capture noise rather than meaningful token patterns.

The $\mathrm{CV}_\mathrm{mean}$ metric shows a consistent decrease as the number of centers increases, dropping from 0.3218 (1 center) to 0.1364 (32 centers).
This indicates that a finer-grained partitioning of the input space leads to more uniform expert utilization. 
Although the 32-center configuration yields the most balanced expert workload (lowest $\mathrm{CV}_\mathrm{mean}$),
the 16-center configuration achieves the optimal balance between maximizing task performance and significantly improving expert utilization compared to configurations with fewer centers.

\begin{table}[h]
    \vspace{-2mm}
    \centering
    \caption{Ablation study on the effect of Component Reactivation Strategy.}
    \vspace{-2mm}
    \label{tab:reactivation_ablation}
    \resizebox{\columnwidth}{!}{
        \begin{tabular}{cc|cccc|c}
            \toprule
            \textbf{Re.act.} & \textbf{Centers} & \textbf{MME} & \textbf{TextVQA} & \textbf{VizWiz} & \textbf{SQA$^\text{I}$} & $\mathbf{\mathrm{CV}}_\mathrm{mean}$\\
            \midrule
            \checkmark & 16 & \textbf{1372} & \textbf{51.86} & \textbf{43.11} & 65.05 & 0.1437 \\
            \ding{55} & 16 &  1360.88 & 51.51 & 41.36 & \textbf{65.54} & 0.1544 \\
            \bottomrule
        \end{tabular}
    }
    \vspace{-3mm}
\end{table}
\paragraph{Impact of component reactivation strategy}
Table \ref{tab:reactivation_ablation} 
analyzes the impact of the Reactivation Strategy for the IDA-MoE model (StableLM 1.6B backbone, 16 centers).
The results show that incorporating Reactivation yields performance improvements on several benchmarks, increasing the MME score by +11.12 points, TextVQA accuracy by +0.35 points, and VizWiz accuracy by +1.75 points.
Although a slight decrease of 0.49 points is observed on SQA$^\text{I}$, the strategy demonstrates a generally positive impact on model performance across the evaluated datasets.
Furthermore, as the Reactivation Strategy is designed to accelerate the convergence of input distribution modeling, we can see a further improved $\mathrm{CV}_\mathrm{mean}$ value from 0.1544 to 0.1437.
These findings suggest the Reactivation Strategy's overall positive contribution to both task performance and expert load balancing for MoE routing.

\subsection{Quantitative Analysis}

\begin{figure}[h]
    \centering
    \begin{subfigure}{0.49\linewidth}
        \centering
        \includegraphics[width=\linewidth]{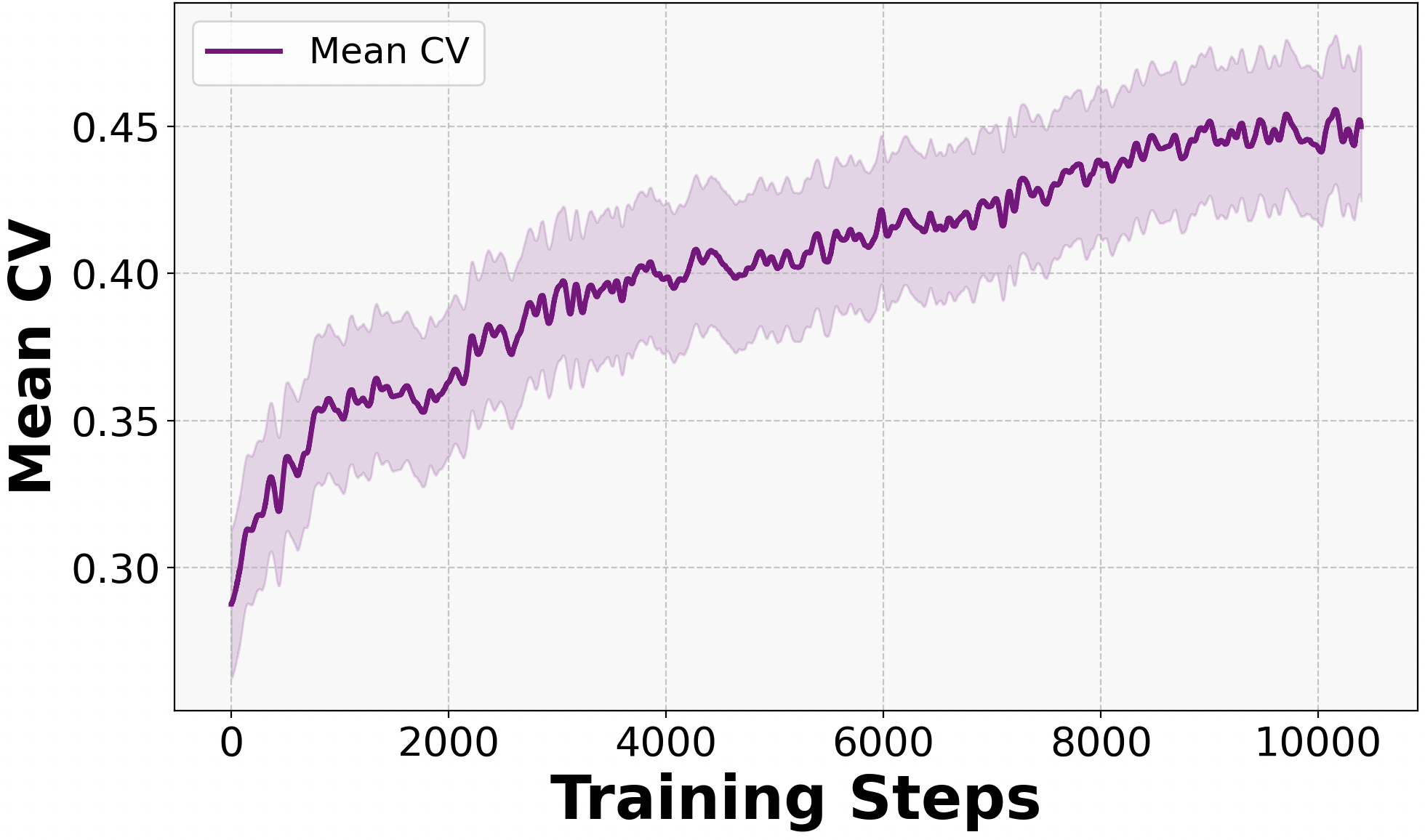}
        \caption{MoE LLaVA without auxiliary loss}
        \label{fig:nlb_cv}
    \end{subfigure}
    \vspace{2mm}
    \begin{subfigure}{0.49\linewidth}
        \centering
        \includegraphics[width=\linewidth]{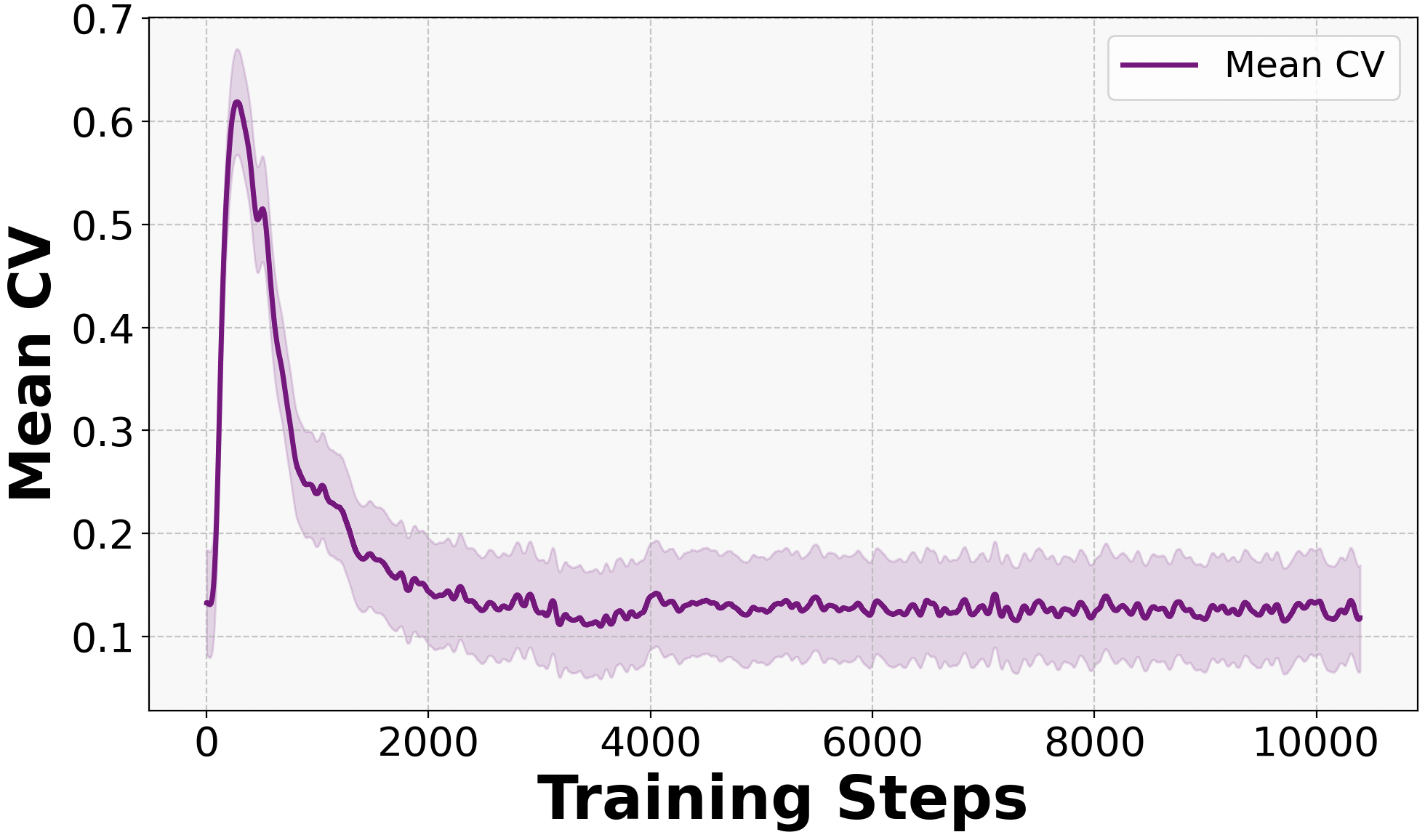}
        \caption{IDA MoE}
        \label{fig:gmm_cv}
    \end{subfigure}
    \vspace{-3mm}
    \caption{$\mathbf{\mathrm{CV}}_\mathrm{mean}$ change over training.}
    \label{fig:training_cv}
    \vspace{-3mm}
\end{figure}


\paragraph{Decoupling Routing From Task-Specific Loss Leads to Natural Load Balance}
As analyzed in \cref{subsec:imbalanced}, conventional routing mechanisms exhibit a persistent tendency toward imbalanced expert utilization.
As illustrated in \cref{fig:nlb_cv}, traditional MoE Architecture without auxiliary loss exhibits steadily increasing $\mathrm{CV}_\mathrm{mean}$ values, indicating progressively worsening load imbalance as training goes. 
In contrast, by decoupling the training of token allocation from task performance, \cref{fig:gmm_cv} demonstrates naturally declining $\mathrm{CV}_\mathrm{mean}$ values throughout the training without introducing explicit auxiliary balancing losses. 
This emergent load-balancing behavior arises from the nature of convergence Gaussian Mixture Modeling of the input space. 
IDA-MoE establishes a more principled foundation for expert specialization while maintaining balanced utilization, all without the need for heuristic regularization terms that can potentially compromise task performance.

\begin{figure}[h]
    \centering
    \begin{subfigure}{0.49\linewidth}
        \centering
        \includegraphics[width=\linewidth]{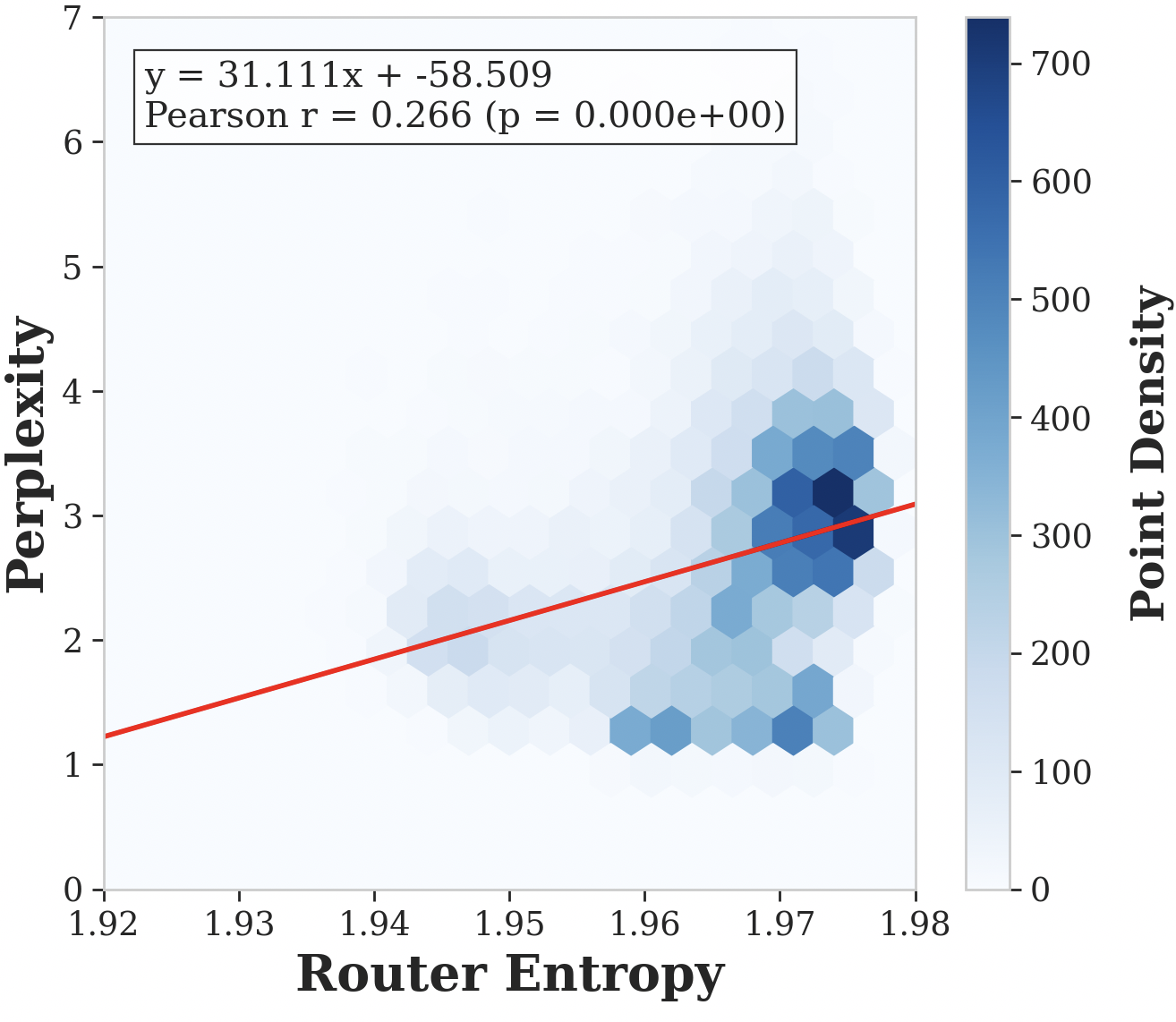}
        \caption{MoE LLaVA}
        \label{fig:lb_ent}
    \end{subfigure}
    \vspace{2mm}
    \begin{subfigure}{0.49\linewidth}
        \centering
        \includegraphics[width=\linewidth]{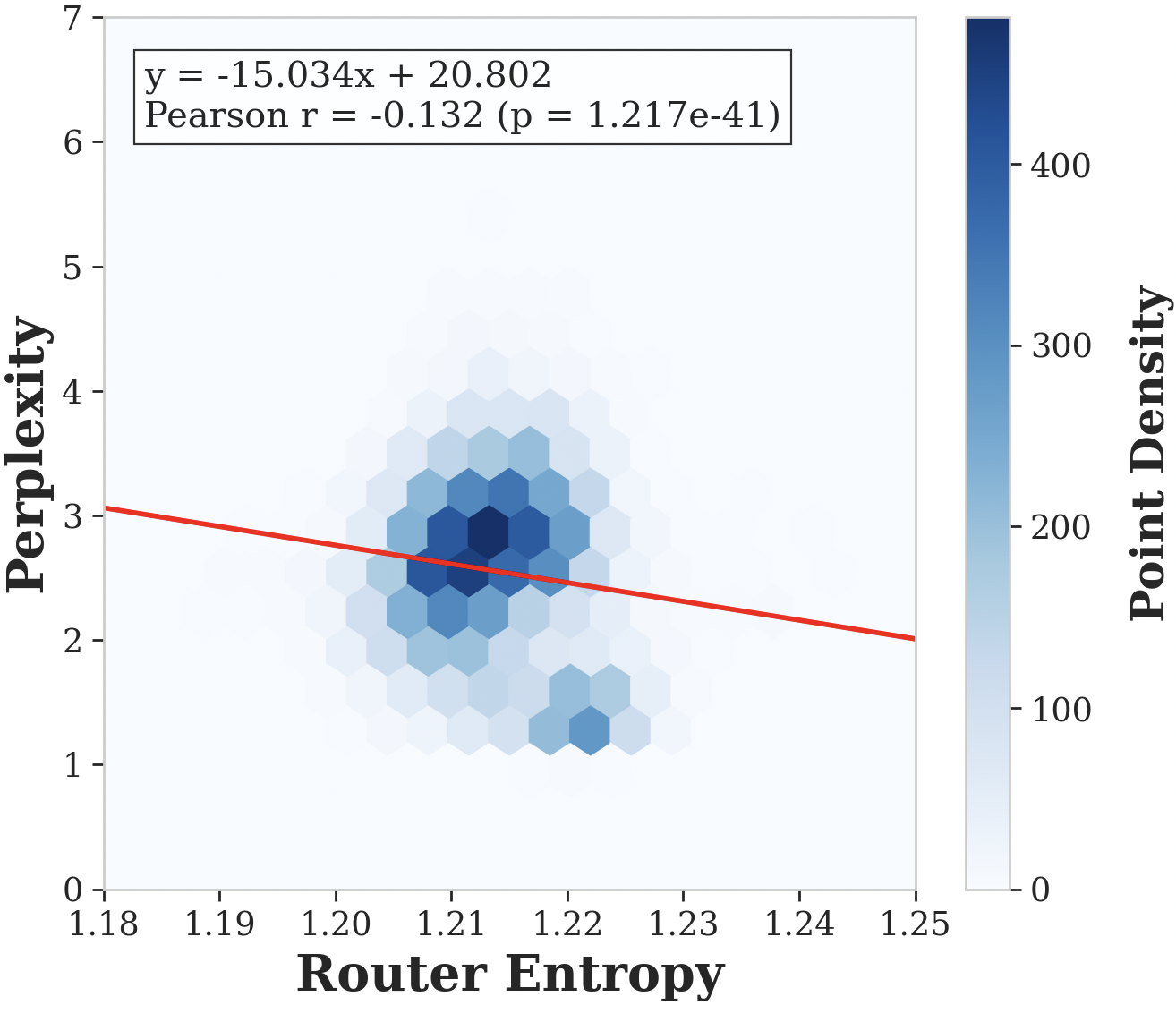}
        \caption{IDA MoE}
        \label{fig:gmm_ent}
    \end{subfigure}
    \vspace{-3mm}
    \caption{\textbf{Routing Entropy \& Perplexity relationship comparison between MoE LLaVA and IDA MoE.} Hexbin density plots with regression lines illustrate how routing decisions affect model performance. X-axis represents routing entropy (lower is more decisive), Y-axis shows perplexity (lower is better performance).}
    \vspace{-6mm}
    \label{fig:ent_perpl}
\end{figure}

\paragraph{Uncertain Routing Decision leads to performance degradation}

As analyzed in \cref{subsec:Specialization}, 
auxiliary loss will cause uncertain token-expert allocation.
During training, it causes instability where token assignments frequently change (see Supplementary Sec. 4 for detailed analysis).
At inference time, this instability makes routing vulnerable to noise and input distribution shifts.
In practical applications, discrepancies between training and test data can result in suboptimal routing decisions and consequently degrade performance.
To quantify this effect, we examine the relationship between routing entropy (a measure of routing uncertainty) and perplexity.
In our analysis, we adopt 4 experts with top-2 activated configuration.
The entropy ranges from 0 to 2, with the ideal value being 1 (representing two decisive experts, each with 0.5 gating weight).
Values closer to 0 indicate one dominant expert, while values approaching 2 suggest indecisive routing across all experts.

As shown in \cref{fig:lb_ent}, MoE-LLaVA's routing entropy typically ranges from 1.96 to 1.98, indicating highly uncertain routing decisions.
We observe a positive correlation between routing entropy and perplexity (Pearson coefficient 0.266), suggesting that indecisive routing contributes to degraded performance.
In contrast, IDA-MoE's routing entropy typically falls between 1.20 and 1.23, much closer to the optimal value of 1, indicating clearer routing decisions.
Notably, IDA-MoE exhibits a slight negative correlation (-0.132) between routing entropy and perplexity.
The negative correlation in IDA-MoE likely stems from slight overfitting on some over-simplified input clusters.
\vspace{-2mm}
\section{Conclusion}
In this paper, we introduce Input Domain Aware Mixture of Experts (IDA-MoE), a novel approach that fundamentally rethinks expert routing in vision language models. By decoupling the router training from task-specific loss functions and employing a principled probabilistic model for token allocation, IDA-MoE 
naturally strikes a balance between load distribution and model performance without relying on auxiliary losses or heuristic constraints. Empirical results demonstrate that IDA-MoE consistently outperforms previous routing and load balancing mechanisms across a diverse set of benchmarks. This performance advantage stems from more decisive routing decisions and better-specialized experts, while maintaining excellent load balance.

\begin{acks}
This research was supported by the National Natural Science Foundation of China (Grant No. 62276245).
\end{acks}

\bibliographystyle{ACM-Reference-Format}
\bibliography{reference}

\end{document}